\documentclass{article}
\usepackage{spconf,amsmath,graphicx}
\usepackage{multirow}
\usepackage{makecell}

\title{Multistream neural architectures for Cued Speech recognition using a pre-trained visual feature extractor and constrained CTC decoding}
 
\name{Sanjana Sankar, Denis Beautemps, and Thomas Hueber \thanks{This work, as part of the Comm4CHILD project, has received funding from the European Union’s Horizon 2020 research and innovation programme under the Marie Sklodowska-Curie Grant Agreement No 860755. Authors would like to thank Sylla Camara for fruitful discussions.}}
\address{Univ. Grenoble Alpes, CNRS, Grenoble INP, GIPSA-lab, 38000 Grenoble, France }
\begin{document}
%
\maketitle
\begin{abstract}
This paper proposes a simple and effective approach for automatic recognition of Cued Speech (CS), a visual communication tool that helps people with hearing impairment to understand spoken language with the help of hand gestures that can uniquely identify the uttered phonemes in complement to lip-reading. The proposed approach is based on a pre-trained hand and lips tracker used for visual feature extraction and a phonetic decoder based on a multistream recurrent neural network trained with connectionist temporal classification loss and combined with a pronunciation lexicon.  The proposed system is evaluated on an updated version of the French CS dataset CSF18 for which the phonetic transcription has been manually checked and corrected. With a decoding accuracy at the phonetic level of 70.88\%, the proposed system outperforms our previous CNN-HMM decoder and competes with more complex baselines.

\end{abstract}
\section{Introduction}
\label{sec:intro}
\begin{figure}
    \centering
    \includegraphics[width=8.3cm]{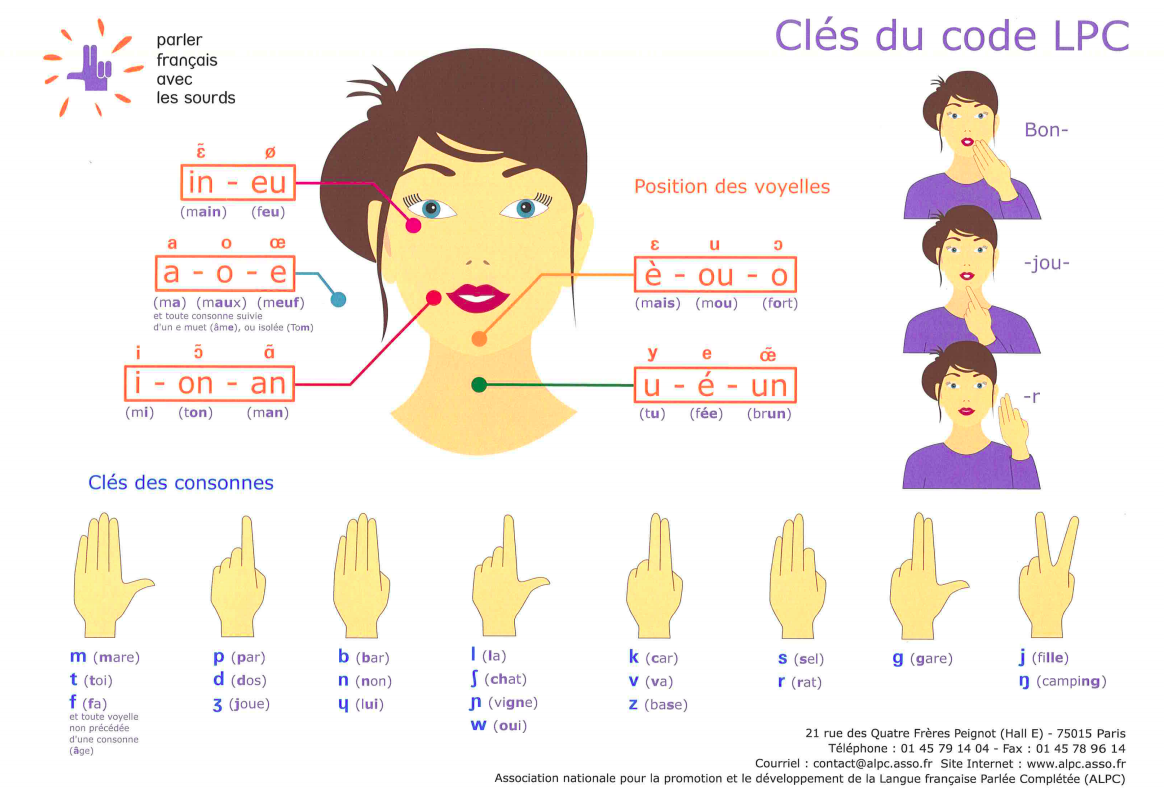}
    \caption{Overview of the \textit{Langue fran{\c c}aise Parl{\'e}e Compl{\'e}t{\'e}e} (Cued-speech system for French)}
    \label{fig:LPC}
\end{figure}

Cued Speech (CS) is a visual communication tool developed by Cornett \cite{R1} in 1967 to help people with hearing impairment to better understand the spoken language. It encodes speech as a combination of visible hand shapes (for consonants) and hand positions (for vowels) to highlight the uttered phoneme and complement lip-reading \cite{R2}. French CS or \textit{Langue fran{\c c}aise Parl{\'e}e Compl{\'e}t{\'e}e (LPC)}  \cite{R3} uses five hand positions to encode vowels and eight hand shapes for consonants as shown in Fig \ref{fig:LPC}.

Automatic Cued Speech Recognition (ACSR) is based on transcripting visual cues of speech to text. The first ACSR systems were focused on isolated vowel and/or consonant recognition \cite{R4}. The cuers were artificially marked with hand landmarks and lip makeup before the video recording to simplify the image processing \cite{R4}. Currently, the research is focused on continuous ACSR, i.e., the decoding of connected words \cite{R5}.
In \cite{R6}, we proposed the first system for continuous phoneme decoding from CS visual data. A feature extractor based on a multi-stream convolutional neural network (CNN) processing the raw regions of interest of hand and lips was combined with an HMM-GMM phonetic decoder. One of the major challenges in continuous ACSR is to deal with the asynchrony between hand and lip \cite{R7}, i.e., the configuration of the hand pertaining to a certain phoneme can precede (or follow) that of the lips for the same phoneme by a variable delay ranging from a few milliseconds to several hundred milliseconds \cite{R15}. In \cite{R6}, this issue has been  addressed with a simple heuristic by considering the hand configuration observed at the beginning of the previous phoneme. However, this  requires a temporal segmentation of the visual streams at the phonetic level. One of the goal of the present study is to get rid of this time-consuming task. To that purpose, we investigated the use of recurrent neural networks (Bi-directional Gated Recurrent Units (Bi-GRUs) \cite{R11}) trained with a Connectionist Temporal Classification (CTC) loss  \cite{R12}. In particular, we compare different architectures for combining lips and hand information within the network. 

The second major challenge in continuous ACSR is the lack of datasets. To the best of our knowledge, the largest available corpus of CS data is the CSF18 corpus released with our previous study \cite{R6}. Since this corpus remains relatively small (i.e. 476 sentences), we hypothesize that training the feature extractor from scratch (i.e. a CNN  in our previous study) may not be an efficient strategy. Therefore, we investigate in the present study the use of a pre-trained feature extractor. We use the Mediapipe \cite{R16} toolkit to infer automatically a set of landmarks of the hand and lips from a raw image sequence of a CS cuer. 

Since our previous study \cite{R6}, other approaches have been proposed for decoding continuous CS data at the phonetic level. In \cite{R8}, Papatimitriou et al. proposed a fully convolutional model with a time-depth separable block and attention based decoder. In \cite{R9}, Wang et al. make use of pre-trained teacher model by exploring different methods for cross-modal knowledge distillation. Both approaches are evaluated on the CSF18 corpus and outperformed the CNN-HMM baseline by a significant margin. 
However, we identified a potential bias in some of those studies due to the particular structure of the linguistic material of the CSF18 corpus. One of the goal of this paper is thus to quantify and report the impact of this potential bias on the overall performance. Moreover, we also show that a performance gain could be obtained by correcting the phonetic transcription of the CSF18 corpus so that it reflects what the CS cuer has actually coded. 

Finally, all previous studies in ACSR focus on phonetic decoding. However, a practical system should be able to deliver to the user the most likely sequence of words. As a first step toward such a system, we also investigated the use of the Token Passing algorithm combined with a  pronunciation lexicon. The key contributions of this work are (i) a light architecture for continuous decoding of CS based on a pre-trained feature extractor and a multistream RNN which does not need a temporal segmentation of the visual stream at the phonetic level and can compete with the more complex approaches recently reported in the literature, (ii) the release of a corrected version of the CSF18 which now reflects more accurately the CS encoded keys and can lead to a significant performance gain, and (iii) a first attempt toward automatic decoding of CS at the word level.

\section{Methodology}


Visual features are extracted automatically with the use of Mediapipe toolkit. The key regions of interest for feature extraction in an ACSR system are the hands and the lips. MediaPipe Hands and MediaPipe Face Mesh provide pre-trained solutions that can estimate the 3D geometry of the visuals. These solutions were adapted for the purpose of this experiment. MediaPipe Hands is a high-fidelity hand and finger tracking solution that provides 21 3D landmarks of a hand. MediaPipe Face Mesh is a face geometry solution that estimates 468 3D face landmarks. For feature extraction, the raw input image sequences for 476 French sentences of the CSF18 corpus were fed to the system. For the purpose of CS, only 42 2D landmarks that attribute to the lips and 21 2D landmarks on the hands were considered for this study. The 2D landmarks of hands and lips pertain to the x and y coordinates of their position in the image. A Principal Component Analysis (PCA) \cite{R13} of the x-y coordinates was done and normalized for each of the two streams dedicated to hand and lips. The first 20 principal components that can summarize up to 99\% of the information contained in the features are selected for each stream in order to reduce the data taxation on the model. The finger tracking for the index or the middle finger was also done separately to retain the information for hand position which encodes the vowel. Since in French CS, either the index finger or the middle finger is always seen for all the target gestures, the extremity of these fingers are tracked and fed to the 3rd input stream in Fig \ref{fig:model} (b). This is done in order to accentuate the information for vowel decoding.

\subsection{Model Architecture}
\label{ssec:acsr}

\begin{figure*}[htb]
\begin{minipage}[b]{.38\linewidth}
  \centering
  \centerline{\includegraphics[width=0.9\textwidth]{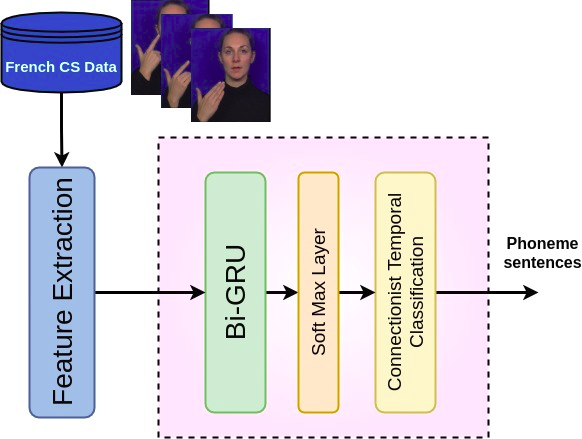}}
  \vspace{0.2cm}
  \centerline{(a) Early Fusion ACSR}\medskip
\end{minipage}
\hfill
\begin{minipage}[b]{0.60\linewidth}
  \centering
  \centerline{\includegraphics[width=0.9\textwidth]{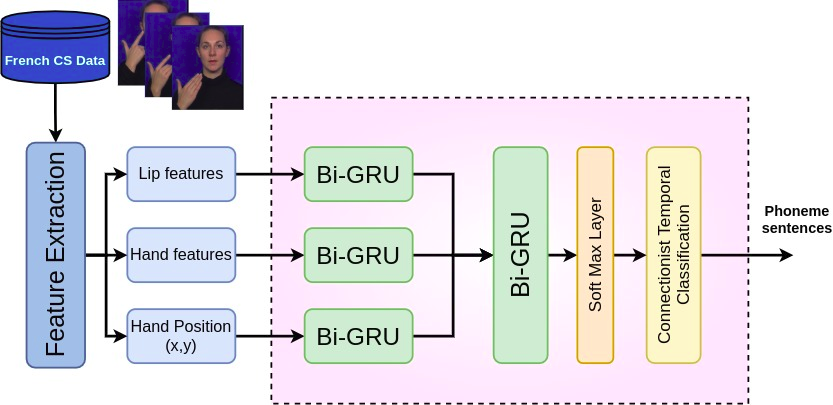}}
  \centerline{(b) 3-stream ACSR system}\medskip
\end{minipage}
\caption{Proposed architectures for automatic Cues Speech Recognition}
\label{fig:model}
\end{figure*}

In this study, we use Bi-GRUs to capture the intrinsic dynamics of hand and lips in CS. Three architectures are investigated: \textbf{(A)}  Early Fusion ACSR, Fig.\ref{fig:model}(a), \textbf{(B)} 2-Streams (only the Lips and Hand streams in Fig.\ref{fig:model}(b)), and \textbf{(C)} 3-Streams ACSR, Fig.\ref{fig:model}(b). In \textbf{(B)} and \textbf{(C)}, we employ middle fusion. The extracted features for each stream are fed to a Bi-GRU that caters specifically to each stream. The output of the stream-wise GRUs are then concatenated and fed to a second layer of Bi-GRU as shown in Fig. \ref{fig:model}(b). The purpose of including the second Bi-GRU layer after concatenation in \textbf{(B)} and \textbf{(C)} is to intrinsically learn the dynamics between each stream. Bi-GRUs can capture both past and future information. Therefore, we employ Bi-GRU to learn long-term dependencies between various time steps of the different streams. This step is crucial to automatically learn the time lag between the hand gesture and lip movement in CS.

A fully connected softmax layer then provides the posterior distribution for each phonetic class. This network is trained with a CTC loss which decodes directly the most possible sequence of phonemes without the need of temporal segmentation of the visual input data.

\subsection{Decoding Strategies}
\label{ssec:wordrec}
Two decoding strategies were investigated. In the first one, referred to as "unconstrained", a greedy CTC decoder is used to recover the most likely phonetic sequence. In short, the decoder first concatenates the most probable phonemes at each timestep and then removes the duplicate phonemes and the CTC blank tokens. The second strategy, hereafter the "constrained" one, exploits a pronunciation lexicon and is based on the Token Passing algorithm \cite{young1989token}, adapted for CTC \cite{R17}. The use of a lexicon allows us (i) to introduce prior linguistic knowledge in the decoding and thus potentially resolve ambiguities in the CS data, (ii) to recover a sequence of words which is a crucial step toward a practical system of AVCSR.  

\section{Experiment}
\label{sec:pagestyle}

\subsection{Dataset}
\label{ssec:dataset}
The CSF18 \cite{R6, R10} is a dataset consisting of 238 French sentences that has been uttered twice. The French CS cuer was recorded at 50 fps and the image is of $720$ X $576$ pixels resolution. We conducted experiments on the publicly available CSF18 dataset. This corpus constitutes of 34 phonetic classes - 14 vowels and 20 consonants that can be encoded with 8 hand shapes and 5 hand positions by the cuer. We then updated this corpus by manually correcting the phonetic transcription so that it matches the CS keys encoded by the cuer, wherever it differed from the phonemes transcribed based on speech. In this process, 3 new phonetic classes were introduced in the corpus- the clusters \textit{"gn"}, \textit{"ng"} and the semi-vowel \textit{"ui"} as shown in Fig.\ref{fig:LPC}. This helps in the correct labeling of the gestures and hand positions used by the cuer. The updated corpus is henceforth referred to as CSF18V2 and is made  available with the DOI: 10.5281/zenodo.1206000).


\subsection{Model Training}
The first ACSR model shown in Fig.\ref{fig:model}(a) consists of a single layer of Bi-GRU with 128 fully connected hidden units. The 2-Streams and 3-Streams ACSR models consist of 1-layer Bi-GRU for each stream and the second layer of Bi-GRU after concatenation as shown in Fig.\ref{fig:model}(b) constitutes of 256 hidden units. The model is trained with Adam Optimizer \cite{R14} with an initial learning rate of $0.001$ and decreased by a factor of $2$ when necessary. The training patience is set to $10$ and the CTC Loss is used to calculate the loss between all the possible phonetic transcriptions at train time. The mini-batch size is fixed to $16$. We have setup 3 different experiments and all the models are trained using the same optimization criteria and loss function. At test time, and for the constrained decoding strategy only, a pronunciation lexicon was built by keeping the 1105 words of the CSF18 corpus. 

\subsection{Train-Test split}
As mentioned in \ref{sec:intro}, one of the goals of this paper is to quantify a potential bias in the performance due to the linguistic content of the CSF18 corpus. Since this corpus consists of repeated sentences, a conventional shuffling of the data is very likely to create a large linguistic overlap between training and test set. In another words, a significant number of test and train sentences will share the same text. Neglecting this potential risk can bring a significant difference in the accuracy rate, thus biasing the evaluation of the method. In this study, we carefully control this aspect. First, we kept the order of the sentences unchanged and use $90\%$ of the dataset for training and the remaining $10\%$ for testing. Then, k-fold cross-validation was performed to obtain the average performance over all possible splits without any overlap. Second, we voluntarily shuffle the dataset before splitting into training and test set. This allows us to quantify the potential bias related to the linguistic overlap between training and test sets.  

\begin{table*}[!ht]
\centering
 \begin{tabular}{| c | c | c | c | c | c |} 
 \hline
 Study & Features & Corpus & Model & Shuffling & Acc.(\%) \\ [0.5ex] 
 \hline\hline
 Liu et al. \cite{R6} &  CNN &  & HMM-GMM & No & 61.5\\
 \cline{1-2}\cline{4-6}
 Papadimitriou et al. \cite{R8} &  CNN & CSF18 & Fully Conv. & Yes & 70.9 \\
 \cline{1-2}\cline{4-6}
 Wang et al. \cite{R9} & CNN & &Cross-distillation & No & 74.2\\
 \hline
 \multirow{7}{*}{\textbf{Ours}} & & CSF18 V2 &Early Fusion &  No & 58.8 \\
                        \cline{3-6}
                        & & CSF18 V2 &2-streams Bi-GRUs (unconstrained)&  No & 63.5 \\
                        \cline{3-6}
                        &  & CSF18 &3-streams Bi-GRUs (unconstrained)&  No & 64.9 \\
                        \cline{3-6}
                        & \textbf{Mediapipe} & CSF18 &3-streams Bi-GRUs (unconstrained)&  Yes & 77.1 \\
                        \cline{3-6}
                        &  & & \textbf{3-Streams Bi-GRUs (unconstrained)} &  \textbf{No} & \textbf{70.9} \\
                        \cline{4-6}
                        & & \textbf{CSF18 V2} & 3-Streams Bi-GRUs (unconstrained) & Yes & 79.2 \\
                        \cline{4-6}
                        & & &\textbf{3-Streams Bi-GRUs (constrained)} & \textbf{No} &\textbf{ 70.2} \\
 \hline
 \end{tabular}
 
 \caption{\label{tab:comparison} Results comparison for existing models and ours with and without shuffling train-test split which can bias the evaluated results. Early Fusion uses Fig.\ref{fig:model}(a) architecture, 2-Streams is Fig.\ref{fig:model}(b) with only Lips and Hand streams, and 3-Streams (also constrained) is the same as Fig.\ref{fig:model}(b). For all experiments, $\Delta_{95\%}$ confidence interval is around 4\%.  The results for the other approaches are reported as is from the papers \cite{R6},\cite{R8},\cite{R9}.}
\end{table*}

\subsection{Results and Discussion}
\label{ssec:results}
The evaluated results are expressed in terms of Accuracy $Acc. (\%)= (N-D-S-I) /\/ N$, where $N$, $D$, $S$, $I$ are the number of phonemes in the test set, deletions, substitutions and insertions. The statistical significance of this measurement was assessed by calculating the Binomial proportion confidence interval $\Delta_{95\%}$ using the Wilson formula. 

Results of all conducted experiments are presented in Table \ref{tab:comparison}. 
First, correcting the phonetic transcription of the CSF18 corpus led to a performance gain of $6\%$ (64.9\% vs. 70.9\% when considering 3-streams and unconstrained decoding). In addition, we report, here, the minimum accuracy (68.52\%), the average accuracy (70.9\%) and the best accuracy (72.33\%) obtained across 10 splits with the 3-streams architecture on the CSF18-V2 corpus. This corpus should now be a reference corpus for future studies in the field. 

We now compare the different architectures used to combine hand and lip information within our model. At first, and similar to our previous study, the early fusion strategy provided the the worst results (58.8\%). The 3-streams approach outperforms significantly the 2-streams one (e.g. 70.9\% vs. 63.5\%). Although the hand features include information about the position of all the points in hand, re-enforcing the hand position features by providing the third stream with index/middle finger tracking helps the model further. This highlights the importance of hand position in CS, an expected outcome as the hand position encodes the vowel information. The success of the middle fusion models provide reason to believe that the advancement of the hand with respect to the lips and its variability are learnt by the system.

Our results also highlight the bias due to the dataset shuffling while partitioning it into training and test subsets. This apparently harmless procedure can lead to significant gain of performance of more than  $8\%$ (70.9\% vs. 79.2\% and 64.9\% vs. 77.1\%). This bias should be considered carefully when comparing the different studies in the field. 

We now discuss the performance of the proposed approach w.r.t other studies. First, the proposed method outperforms our previous CNN-HMM approach \cite{R6} by a margin of 4.8\% (60.1\% vs. 64.9\%). Let us recall that the proposed method does not rely on the temporal segmentation of the visual data at the phonetic level. Then, to compare our approach with the method reported in the \cite{R8}, we have to consider the biased condition due to dataset shuffling (as it seems to be the case in this study, see section IV-A in \cite{R8}) and the experiments conducted on the original CSF18 corpus. For these particular experimental conditions, our proposed method outperforms the one reported in \cite{R8} (77.1\% vs. 70.9\% \cite{R8}). 
However, when considering the original CSF corpus, our system performs significantly worse than the one proposed in \cite{R9} (64.9\% vs. 74.2\%) which is based on a more complex architecture (pre-trained teacher model with knowledge distillation toward a student model). 
When considering the corrected CSF18V2 corpus, our method reaches a comparable level of performance (70.9\%). Thus, it would be interesting for future work to quantify the extra benefit given by knowledge distillation approach of \cite{R9} on this updated dataset.  

Finally, we discuss the performance w.r.t to the CTC decoding strategy. First, the use of a $\sim$1000 words pronunciation lexicon (constrained decoding) is not as helpful as expected (no significant difference observed between the two decoding strategies, i.e. 70.9\% vs. 70.2\%). Moreover, the performance at the word level remains very low  ($\sim$25\% correctness). Nevertheless, some sentences were perfectly re-transcribed at the word level and others were partially correctly transcribed, e.g. "\textit{le bedeau euphorique secoue l'anneau un jour par an}" decoded as "\textit{le bedeau euphorique se qu' l' nos un jours part en}".  These results may suggest that accurate decoding of CS at the word level would be possible after integrating  more prior linguistic knowledge via the use of a statistical language model.  

\section{Conclusion and Perspectives}
\label{sec:conclusion}
To conclude, this study has explored and proposed a light model for ACSR which outperforms our previous approach and compete with other recent (and more complex) approaches. The use of an efficient pre-trained feature extractor allowed to reduce the complexity of the (visual) phonetic decoder. Also, cleaning the CSF18 dataset to account for differences in cueing and pronunciation of certain phonemes improved significantly the performance. However, the decoding performance at the word level, as well as the introduction of prior linguistic knowledge via the use of a pronunciation lexicon were not as helpful as expected. Recent neural language models, and the use of larger and multi-speaker datasets could lead to significant improvements in decoding performance.   


\vfill\pagebreak


\bibliographystyle{IEEEbib}
\bibliography{refs}

\end{document}